\documentclass[10pt,twocolumn,letterpaper]{article}

\usepackage{wacv}
\usepackage{times}
\usepackage{epsfig}
\usepackage{graphicx}
\usepackage{amsmath}
\usepackage{amssymb}
\usepackage{caption}
\usepackage{array,multirow,booktabs}
\usepackage{hyperref}
\usepackage{capt-of,etoolbox}
\usepackage{amsfonts}

\wacvfinalcopy % *** Uncomment this line for the final submission

\def\wacvPaperID{41} % *** Enter the wacv Paper ID here

\ifwacvfinal\fi
\begin{document}

%%%%%%%%% TITLE
\title{A deep learning pipeline for product recognition on store shelves}

% Authors at the same institution
%\author{First Author \hspace{2cm} Second Author \\
%Institution1\\
%{\tt\small firstauthor@i1.org}
%}
% Authors at different institutions
\author{Alessio Tonioni\\
DISI, University of Bologna\\
{\tt\small alessio.tonioni@unibo.it}
\and
Eugenio Serra\\
DISI, University of Bologna\\
{\tt\small eugenio.serra@studio.unibo.it}
\and
Luigi Di Stefano\\
DISI, University of Bologna\\
{\tt\small luigi.distefano@unibo.it}
}

\maketitle
\ifwacvfinal\thispagestyle{empty}\fi

%%%%%%%%% ABSTRACT
\begin{abstract}
Recognition of grocery products in store shelves poses peculiar challenges. Firstly, the task mandates the recognition of an extremely high number of different items, in the order of several thousands for medium-small shops, with many of them featuring small inter and intra class variability. Then, available product databases usually include just one or a few studio-quality images per product (referred to herein as \textbf{reference} images), whilst at test time recognition is performed on pictures displaying a portion of a shelf containing several products and taken in the store by cheap cameras (referred to as \textbf{query} images). Moreover, as the items on sale in a store as well as their appearance change frequently over time, a practical recognition system should handle seamlessly new products/packages. Inspired by recent advances in object detection and image retrieval, we propose to leverage on state of the art object detectors based on deep learning to obtain an initial product-agnostic item detection. Then, we pursue product recognition through a similarity search between global descriptors computed on \textit{reference} and cropped \textit{query} images. To maximize performance, we learn an ad-hoc global descriptor by a CNN trained on  \textit{reference} images based on an image embedding loss. Our system is computationally expensive at training time but can perform recognition rapidly and accurately at test time.
\end{abstract}

%%%%%%%%% BODY TEXT
\section{Introduction}
\label{sec:introduction}
\begin{figure}
	\centering
	\includegraphics[width=0.45\textwidth]{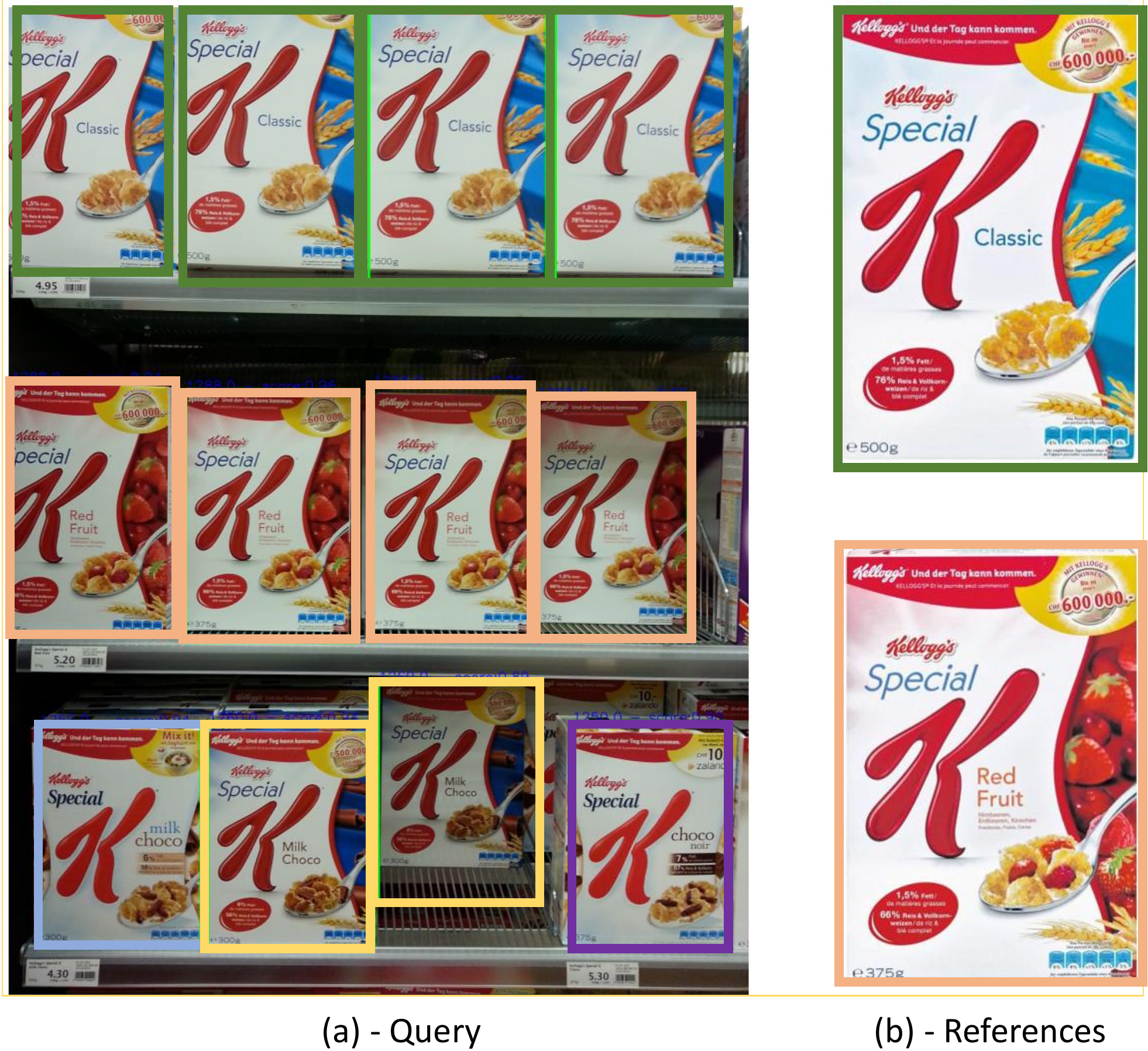}
	\caption{Given a \textit{query} image featuring multiple products (a), our system first detects the regions associated with the individual items and then recognizes the product enclosed in each region based on a database featuring only one \textit{reference} image per product (two examples are shown in (b)). All the products are correctly recognized in (a) with bounding boxes colored according to the recognized classes.}
	\label{fig:teaser}
\end{figure}

Recognizing products displayed on store shelves based on computer vision is gathering ever-increasing attention thanks to the potential for improving the customer's shopping experience (\eg, via augmented reality apps, checkout-free stores, support to the visually impaired \dots) and realizing automatic store management (\eg, automated inventory, on-line shelf monitoring\dots). %However, a series of peculiar challenges render this problem very hard in practice.    

The seminal work on product recognition dates back to \cite{merler2007recognizing}, where Merler \etal highlight the peculiar issues to be addressed in order to achieve a viable approach. First of all, the number of different items to be recognized is huge, in the order of several thousands for small to medium shops, well beyond the usual target for current state-of-the-art image classifiers. Moreover, product recognition can be better described as a hard instance recognition problem, rather than a classification one, as it deals with lots of objects looking remarkably similar but for small details (\eg, different flavors of the same brand of cereals). Then, any practical methodology should rely only on the information available within existing commercial product databases, \ie at most just one high-quality image for each side of the package, either acquired in studio settings or rendered (see \autoref{fig:teaser}-(b)). \textit{Query} images for product recognition are, instead, taken in the store with cheap equipment (\eg, a smart-phone) and featuring many different items displayed on a shelf (see \autoref{fig:teaser}-(a)). Unfortunately, this scenario is far from optimal for state-of-the-art multi-class object detectors based on deep learning \cite{redmon2016yolo9000,huang2016speed,ren2015faster}, which require a large corpus of annotated images as similar as possible to the deployment scenario in order to provide good performance. Even acquiring and manually annotating with product labels a huge dataset of in-store images is not a viable solution due to the products on sale in stores, as well as their appearance, changing frequently over time, which would mandate continuous gathering of annotated in-store images and retraining of the system. Conversely, a practical approach should be trained once and then be able to handle seamlessly new stores, new products and/or new packages of existing products (\eg, seasonal packages).

To tackle the above issues, we address product recognition by a pipeline consisting of three stages. Given a shelf image, we perform first a class-agnostic object detection to extract region proposals enclosing the individual product items. This stage relies on a deep learning object detector trained to localize product items within images taken in the store; we will refer to this network as to the \emph{Detector}. In the second stage, we perform product recognition separately on each of the region proposal provided by the \emph{Detector}. Purposely, we carry out K-NN (K-Nearest Neighbours) similarity search between a global descriptor computed on the extracted region proposal and a database of similar descriptors computed on the \textit{reference} images available in the product database. Rather than deploying a general-purpose global descriptor (\eg, Fisher Vectors \cite{perronnin2010large}), we train a CNN using the \textit{reference} images to learn an image embedding function that maps RGB inputs to n-dimensional global descriptors amenable for product recognition; this second network will be referred to as to the \emph{Embedder}.  Eventually, to help prune out false detections and improve disambiguation between similarly looking products, in the third stage of our pipeline we refine the recognition output by re-ranking the first K proposals delivered by the similarity search. An exemplary output provided by the system is depicted in \autoref{fig:teaser}-(a)).

%It is worth observing how the first two stages of our pipeline remind the structure of a region proposal based CNN detector (\eg \cite{ren2015faster}), though, rather than submitting proposals to a classifier, we perform a similarity search based on a learned embedding space wherein similarly looking images tend be appear close together. 

It is worth pointing out how our approach needs samples of annotated in-store images only to train the product-agnostic \emph{Detector}, which, however, does not require product-specific labels but just bounding boxes drawn around items. In \autoref{sec:experimental} we will show how the product-agnostic \emph{Detector} can be trained once and for all so to achieve remarkable performance across different stores despite changes in shelves disposition and product appearance.  
Therefore, new items/packages are handled seamlessly by our system simply by adding their global descriptors (computed through the \textit{Embedder}) in the \textit{reference} database. Besides, our system scales easily to the recognition of thousands of different items, as we use just one (or few) \textit{reference} images per product, each encoded into a global descriptor in the order of one thousand float numbers.  
%Assuming, \eg, to address recognition of  10000 products, recognizing a \textit{query} region proposal boils down to computing its global descriptor and comparing it with 10000 reference ones, an operation that can be carried out in a few hundreds of milliseconds. 

Finally, while computationally expensive at training time, our system turns out light (\ie memory efficient)  and fast at deployment time, thereby enabling near real-time operation. Speed and memory efficiency do not come at a price in performance, as our system compares favorably with respect to previous work on the standard benchmark dataset for product recognition.

\section{Related Work}
\label{sec:related}
Grocery products recognition was firstly investigated in the already mentioned paper by Merler \etal  \cite{merler2007recognizing}. Together with a thoughtful analysis of the problem, the authors propose a system based on local invariant features. However, their experiments report performance far from conducive to real-world deployment in terms of   accuracy and speed. A number of more recent works tried to improve product recognition by leveraging on: a) stronger features followed by classification \cite{cotter2014hardware}, b) the statistical correlation between nearby products on shelves \cite{advani2015visual,baz2016context} c) additional information on the expected product disposition \cite{tonioni2017product} or d) a hierarchical multi-stage recognition pipeline \cite{franco2017grocery}. Yet, all these recent papers focus on a relatively small-scale problem, \ie recognition of a few hundreds different items at most, whilst usually several thousands products are on sale in a real shop. George \etal \cite{george2014recognizing} address more realistic settings and propose a multi-stage system capable of recognizing $\sim3400$ different products based on a single model image per product. The authors' contribution includes releasing the dataset employed in their experiments, which we will use in our evaluation. More recently, \cite{yoruk2016efficient} has tackled the problem using a standard local feature based recognition pipeline and an optimized Hough Transform to detect multiple object instances and filter out inconsistent matches, which brings in a slight performance improvement.  

Nowadays, CNN-based systems are dominating object detection benchmarks and can be subdivided into two main families of algorithms based on the number of stages required to perform detection. On one hand, we have the slower but more accurate two stage detectors \cite{ren2015faster}, which decompose object detection into a region proposal followed by an independent classification for each region. On the other hand, fast one stage approaches  \cite{redmon2016yolo9000,huang2016speed} can perform detection and classification jointly. A very recent work has also addressed the specific domain of grocery products, so as to propose an ad hoc detector \cite{qiao2017scalenet} that analyzes the image at multiple scales to produce more meaningful region proposals.

Besides, deploying CNNs to obtain rich image representations is an established approach to pursue image search, both as a strong off-the-shelf baseline \cite{sharif2014cnn} and as a key component within more complex pipelines \cite{gordo2016deep}. Inspiration for our product recognition approach came from  \cite{schroff2015facenet,bell2015learning}. In \cite{schroff2015facenet}, Schroff \etal train a CNN using triplets of samples to create an embedding for face recognition and clustering while in \cite{bell2015learning} Bell \etal rely on a CNN to learn an image embedding to recognize the similarity between design products. Similarly, in the related field of fashion items recognition, relying on learned global descriptor rather than classifiers is an established solution shared among many recent works \cite{hadi2015buy,wang2014learning,shankar2017deep}.  

%eventualmente aggiungere fashion to street
%Our approach reminds the structure of a two-stage object detector though, rather that submitting region proposals to a classifier, we perform a similarity search based on a representation learned by triplets of samples so as to map similarly looking products close one to another into the embedding space.
% As already mentioned in \autoref{sec:introduction}, casting recognition as an image search is key to handle seamlessly thousands of different items as well as new products and packages. As we will show in \autoref{sec:experimental},  a representation learned ad-hoc to tell apart grocery products does yield significant performance improvements with respect to the adoption of a general-purpose descriptor.  

\section{Proposed Approach}
\label{sec:product_detection_recognition}
\begin{figure*}
	\centering
	\includegraphics[width=0.75\textwidth]{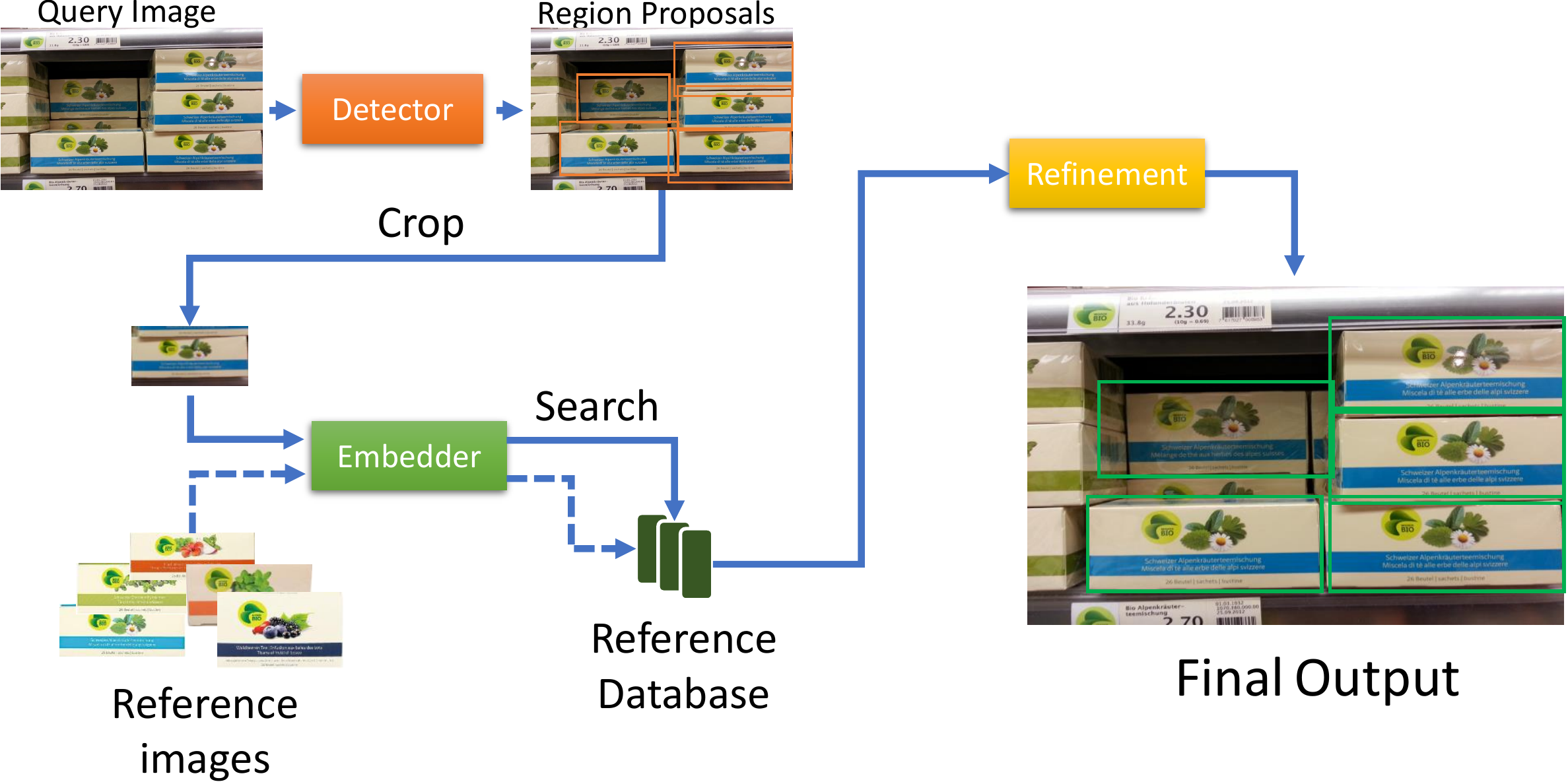}
	\caption{Schematic structure of our proposed product recognition pipeline. Dashed arrows denotes elaboration that can be performed offline just once since are not related to the \textit{query} images.}
	\label{fig:pipeline}
\end{figure*}

\autoref{fig:pipeline} shows an overview of our proposed pipeline. In the first step, described \autoref{ssec:detection}, a CNN (\emph{Detector}) extracts region proposals from the \textit{query} image. Then, as detailed in \autoref{ssec:recognition}, each region proposal is cropped from the \textit{query} image and sent to another CNN (\emph{Embedder}) which computes an ad-hoc image representation. These will then be deployed to pursue product recognition through a K-NN similarity search in a database of representations pre-computed off-line by the \emph{Embedder} on the \textit{reference} images. Finally, as illustrated in \autoref{ssec:refinement}, we combine different strategies to perform a final refinement step which helps to prune out false detections and disambiguate among similar products.

%We perform product detection by an initial region proposal stage followed by product recognition by means of K-NN similarity search between the global descriptors extracted from the regions and those computed on the reference product images. Finally, an optional refinement step may be carried out to re-rank NNs and remove false detections. We use a CNN-based object detector for the first stage and another CNN trained to learn an image embedding for the second. For the final refinement, 

\subsection{Detection}
\label{ssec:detection}
Given a \textit{query} image featuring several items displayed in a store shelf, the first stage of our pipeline aims at obtaining a set of bounding boxes to be used as region proposals in the following recognition stage. Ideally, each bounding box should contain exactly one product, fit tightly the visible package and provide a confidence score measuring how much the detection should be trusted.

State-of-the-art CNN-based object detectors may fulfill the above requirements for the product recognition scenario, as demonstrated in \cite{qiao2017scalenet}. Given an input image, these networks can output several accurate bounding boxes, each endowed by a confidence and a class prediction. To train CNN-based object detectors, such as \cite{redmon2016yolo9000,huang2016speed,ren2015faster}, a large set of images annotated with the position of the objects alongside with their class labels is needed. However, due to the ever-changing nature of the items sold in stores, we do not train the \emph{Detector} to perform predictions at the fine-grained class level (\ie, at the level of the individual products), but to carry out a  product-agnostic item detection. Accordingly, the in-store training images for our \emph{Detector} can be annotated for training just by drawing bounding-boxes around items without specifying the actual product label associated with each bounding-box. This formulation makes the creation of a suitable training set and the training itself easier and faster. Moreover, since the \emph{Detector} is trained only to recognize \textit{generic products} from everything else it is general enough to be deployable across different stores and products.
%This not only renders the training process easier and faster but, and much more importantly, forces the \emph{Detector} to be invariant to the different appearances of the individual items, making it possible to deploy the very same trained network across different stores. 
Conversely, training a CNN to directly predict bounding boxes as well as product labels would require a much more expensive and slow image annotation process which should be carried out again and again to keep up with changes of the products/packages to be recognized. This continuous re-training of the \emph{Detector} is just not feasible in any practical settings. 

%Our \emph{Detector} network is trained to be a generic object detector and produce as outputs region candidates for the following recognition step. 

\subsection{Recognition}
\label{ssec:recognition}
Starting from the candidate regions delivered by the \emph{Detector},  we perform recognition  by means of K-NN similarity search between a global descriptor computed on each candidate region and a database of similar descriptors (one for each product) pre-computed off-line on the \textit{reference} images. Recent works (\eg, \cite{sharif2014cnn}) have shown that the activations sampled from layers of pre-trained CNNs can be used as high quality global image descriptors. \cite{wang2014learning} extended this idea by proposing to train a CNN (\ie, the \emph{Embedder}) to learn a function $E:\mathcal{I}\rightarrow\mathcal{D}$ that maps an input image $i \in I$ into a k-dimensional descriptor $d^k \in \mathcal{D}$ amenable to recognition through K-NN search. Given a set of images with associated class labels, the training is performed sampling triplets of different images, referred to as \emph{anchor} ($i_a$), \emph{positive} ($i_p$) and \emph{negative} ($i_n$),  such that $i_a$ and $i_p$ depict the same class while $i_n$ belongs to a different one. Given a distance function in the descriptor space, $d(\mathbf{X},\mathbf{Y})$, with $X,Y\in\mathcal{D}$, and denoted as $E(i)$ the descriptor computed by the the \emph{Embedder} on image $i$, the network is trained to minimize the so called \emph{triplet ranking loss}: 
\begin{multline}
\label{eq:tripletLoss}
\mathcal{L} = max( 0,d(E(i_a),E(i_p))-d(E(i_a),E(i_n))+\alpha)
\end{multline}
with $\alpha$ a fixed margin to be enforced between the pair of distances. Through minimization of this loss, the network learns to encode into nearby positions within  $\mathcal{D}$ the images depicting items belonging to the same class, whilst keeping items of different classes sufficiently well separated. 

We use the same formulation and cast it for the context of grocery product recognition where different products corresponds to different classes (\eg the two reference images of \autoref{fig:teaser}-(b) corresponds to different classes and could be used as $i_p$ and $i_n$). Unfortunately, we can not sample different images for $i_a$ and $i_p$ due to available commercial datasets featuring just a single exemplar image per product (\ie, per class). Thus, to create the required triplet, at each training iteration we randomly pick two products and use their \textit{reference} images as $i_p$ and $i_n$. Then, we synthesize a new  $i_a$ from $i_p$ by a suitable data augmentation function $A:\mathcal{I}\rightarrow\mathcal{I}$, to make it similar to \textit{query} images (\ie, $i_a=A(i_p)$)\footnote{The details concerning the adopted augmentation function are reported in \autoref{sec:experimental}}.

To perform recognition, firstly, the \emph{Embedder} network is used to describe each available \textit{reference} image $i_r$ by a global descriptor  $E(i_r)$ and thus create the \textit{reference} database of descriptors associated with the products to be recognized. Then, when a \textit{query} image is processed, the same embedding is computed on each of the candidate regions, $i_{pq}$, cropped from the \textit{query} image, $i_q$, so to get $E(i_{pq})$. Finally, for each $i_{pq}$ we compute the distance in the embedding space with respect to each \textit{reference} descriptor, denoted as $d(E(i_{pq}),E(i_{r}))$, in order to sift-out the first $K$-NN of $E(i_{pq})$ in the \textit{reference} database. These are subject to further processing in the final refinement step.

\subsection{Refinement}
\label{ssec:refinement}
The aim of the final refinement is to remove false detections and re-rank the first K-NN found in the previous step in order to fix possible recognition mistakes.

Since the initial ranking is obtained comparing descriptors computed on whole images, a meaningful  re-ranking of the first K-NN may be achieved by looking at peculiar image details that may have been neglected while comparing global descriptors and yet be crucial to differentiate a product from others looking very similar. Thus, both the \textit{Query} and each of the first K-NN \textit{reference} images are described by a set of local features ${F_1,F_2,...,F_k}$,  each consisting in a spatial position $(x_i,y_i)$ within the image and a compact descriptor $f_i$. Given these features, we look for similarities between descriptors extracted from  \textit{query} and \textit{reference} images, to compute a set of matches. Matches are then weighted based on the distance in the descriptor space, $d(f_i,f_j)$ and a geometric consistency criterion relying on the unit-norm vector, $\vec{v}$, from the spatial location of a feature to the image center. In particular, given a match, $M_{ij}=(F^q_i,F^r_j)$, between feature $i$ of the \textit{query} image and feature $j$ of the \textit{reference} image, we compute the following weight: 

\begin{equation}
W_{ij}=\frac{(\vec{v}^{\,q}_i \cdot \vec{v}^{\,r}_j)+1}{d(f^q_i,f^r_j)+\epsilon}
\end{equation}

where $\cdot$ marks scalar products between vectors and $\epsilon$ is a small number to avoid potential division by zero. Intuitivelly $W_{ij}$ is bigger for matching features which share the same relative position with respect to the image center (high $(\vec{v}^{\,q}_i \cdot \vec{v}^{\,r}_j)$) and have descriptors close in the feature space (small $d(f^q_i,f^r_j)$). Finally, the first K-NN are re-ranked according to the sum of the weights $W_{ij}$ computed for the matches between the local features. In \autoref{ssec:ideal_study} we will show how good local features can be obtained at zero computational cost as a by-product of our learned global image descriptor. This refinement technique will be referred to as \textbf{\textit{+lf}}. 

%As modern object detector systems tend to produce a huge number of detections, a simple and fast way of pruning out false ones is to perform \textit{Non Maxima Suppression} (\textit{NMS}) of overlapping predictions according to their confidence value, \ie if two bounding boxes mapped to the same class have an intersection over union greater than a threshold $\tau_{nms}$, we keep only the one featuring the highest confidence score. This strategy could reduce the system recall as the number of detections is reduced, however, those remaining should ideally be more reliable. In the following, we will refer to this refinement technique as \textit{+nms}.

A simple additional refinement step consists in filtering out wrong recognitions by the \emph{distance ratio} criterion \cite{lowe2004distinctive} (\ie, by thresholding the ratio of the distances in feature space between the \textit{query} descriptor and its 1-NN and 2-NN). If the ratio is above a threshold, $\tau_{d}$, the recognition is deemed as ambiguous and discarded. In the following, we will denote this refinement technique as \textbf{\textit{+th}}.

Finally, as commercial product databases typically provide a multilevel classification of the items (\eg, at least instance and category level), we propose a re-ranking and filtering method specific to the grocery domain where, as pointed out by \cite{george2014recognizing}, products belonging to the same macro category are typically displayed close one to another on the shelf.    
%items method that aims to extract the macro category of products present in an image. In fact, if we know this information we can simplify the classification problem by considering only a subset of the complete database.
%To find out this, 
In particular, given the candidate regions extracted from the \textit{query} image and their corresponding sets of K-NN, we consider the 1-NN of the region proposals extracted with a high confidence ($>0.1$) by the \emph{Detector} in order to find the main macro category of the image. Then, in case the majority of detections votes for the same macro category, it is safe to assume that the pictured shelf contains almost exclusively items of that category thus filter the K-NN for all candidate regions accordingly.
%consider the most reliable products (those whose classification distance is less than a given threshold). If the number of “secure” products belonging to the same semantic group exceeds a predefined percentage of the total region proposals then we can think this category to be our target one and, secondly, we can reduce the classification database for the “less-reliable” products (those with a higher value of distance) to a corresponding subset.
%The basic assumption behind this refinement strategy is that, as already pointed out in \cite{george2014recognizing}, products belonging to the same category are usually displayed together in the same portion of a shelf, so if we are able to guess the category of the pictured shelf we can prune out candidates and improve drastically the recognition rate.
It is worth observing how this strategy implicitly leverages on those products easier to identify  (\ie, the high-confidence detections) to increase the chance to correctly recognize the harder ones. We will refer to this refinement strategy as to \textbf{\textit{+mc}}.

\section{Experimental Results}
\label{sec:experimental}
To validate the performance of our product recognition pipeline we take into account two possible use cases dealing with different final users: 
\begin{itemize}
	\item \textbf{Customer}: the system should be deployed for a guided or partially automated shopping experience (\eg, product localization inside the shop or augmented reality overlays or support to visually impaired). As proposed in \cite{george2014recognizing}, in this use case the goal is to detect \emph{at least one} instance of each visible type of products displayed in a shelf picture.
	\item \textbf{Management}: the system will be used to partially automate the management of a shop (\eg, automatic inventory and restocking). Here, the goal is to recognize \emph{all} visible product instances displayed in a shelf picture. 
\end{itemize}
%The following will show how our system can perform fairly well in both scenarios.

\subsection{Datasets and Evaluation Metrics}
\label{ssec:datasets}
\begin{figure}
	\centering
	\begin{tabular}{cc}
		\includegraphics[width=0.22\textwidth]{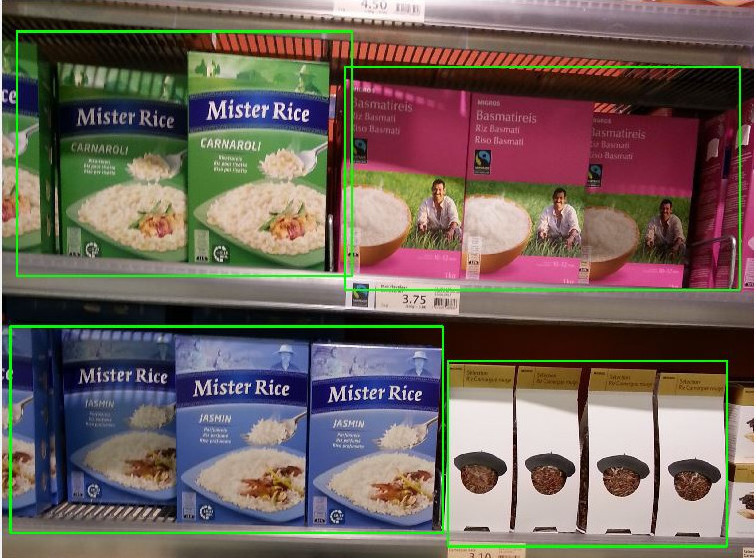} & 
		\includegraphics[width=0.22\textwidth]{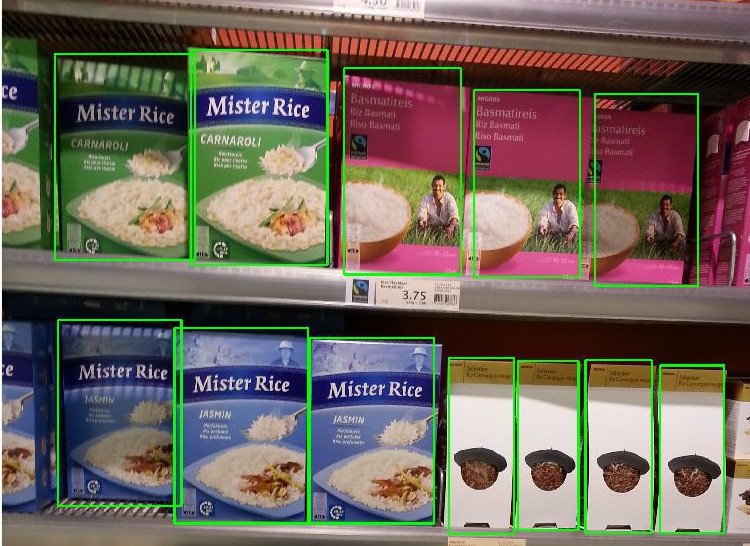} \\
		(a)-Customer\cite{george2014recognizing} & (b)-Management\cite{tonioni2017product}\\
	\end{tabular}
	\caption{In (a) the system should identify at least one instance for each product type, while in (b) it should find and correctly localize all the displayed product instances.}
	\label{fig:bboxes}
\end{figure}
For our experimental evaluation we rely on the publicly available \emph{Grocery Products} dataset  \cite{george2014recognizing}, which features more than 8400 grocery products organized in hierarchical classes and with each product described by exactly one \textit{reference} image. The dataset contains also 680 in-store (\textit{query}) images that display items belonging to the \textit{Food} subclass of the whole dataset. The annotations released by the authors for the \textit{query} images allow for evaluating performance in the \emph{Customer} use case, as they consist in bounding boxes drawn around spatial clusters of instances of the same products. To test our system also in the \emph{Management} use case,  we deploy the annotations released by the authors of \cite{tonioni2017product}, which consist of boxes around each product instance for a subset of 70 in-store pictures of the \emph{Grocery Products} dataset. \autoref{fig:bboxes} shows examples of the two kinds of annotations used to evaluate the system in the two different use cases. 

To compare our work with previously published results we use the metrics proposed by the authors in  \cite{george2014recognizing}: mean average precision-\textit{mAP} (the approximation of the area under the Precision-Recall curve for the detector) and Product Recall-\textit{PR} (average product recall across all the test image). As for scenario (a), we report also mean average multi-label classification accuracy-\textit{mAMCA}.

To train the \emph{Detector} we acquired multiple videos with a tablet mounted on a cart facing the shelves and manually labeled a subset of sampled frames to create a training set of 1247 images. Thus, our videos are acquired in different stores with respect to those depicted in the \emph{Grocery Products} dataset and feature different products on different shelves, vouching for the generalization ability of our system.

\subsection{Implementation Details}
\label{ssec:ideal_study}
For all our tests we have used as \emph{Detector} the state-of-the-art one-stage object detector known as yolo\_v2 (shortened in \textit{yolo}) \cite{redmon2016yolo9000}. We choose this network as it grants real-time performance on a GPU and for the availability of the original implementation. Starting from the publicly available weights\footnote{\url{https://github.com/pjreddie/darknet}}, we have fine tuned the network on our 1247 in-store images for 80000 steps keeping the hyperparameters suggested by the original authors. 

The backbone network for our \textit{Embedder} is a VGG\_16 \cite{Simonyan14c} pre-trained on the Imagenet-1000 classification task (weights publicly available\footnote{\url{https://github.com/tensorflow/models/tree/master/research/slim}}). From this network we obtain global image descriptors by computing \textit{MAC} features \cite{tolias2015particular} on the conv4\_3 convolutional layer and applying L2 normalization to obtain unit-norm embedding vectors. To carry out the comparison between descriptors, both at training and test time, we used as distance function $d(X,Y)=1-X \cdot Y$ with $X,Y \in \mathcal{D}$ (\ie, 1 minus the cosine similarity between the two descriptors). To better highlight the advantage of learning an ad-hoc descriptor in the following we will report experiments using general purpose descriptors obtained without fine tuning the \emph{Embedder} with the suffix $\_gd$ (general descriptor), while  descriptors obtained after fine-tuning as described in \autoref{ssec:recognition} will be denoted by the suffix $\_ld$ (learned descriptor). 
To train the \emph{Embedder} we use the \textit{reference} images of \emph{Grocery Products} dealing with the products belonging to the \textit{Food} subclass (\ie,  3288 different product  with exactly one training image for each). To enrich the dataset and create the anchor images, $i_a$, we randomly perform the following augmentation functions $A$: blur by a Gaussian kernel with random $\sigma$, random crop, random brightness and saturation changes. This augmentations were engineered so to transform the \textit{reference} images in a way that renders them similar to the proposals cropped from the \textit{query} images. 
The hyper-parameters obtained by cross validation for the training process are as follows: $\alpha=0.1$ for the triplet loss, learning rate $lr=0.000001$, ADAM  optimizer and fine-tuning by 10000 steps with batch size 24. 
%Due to the lack of space we refer the readers to the supplementary material for additional details on how we choose this models of networks and which descriptor strategy to use.

We propose a novel kind of local features for the \textit{+lf} refinement: as \textit{MAC} descriptors are obtained by applying a max-pool operation over all the activations of a convolutional layer, by changing the size and stride of the pool operation it is possible to obtain a set of local descriptor with the associated location being the center of the pooled area reprojected into the original image. By leveraging on this intuition, we can obtain in a single forward computation both a global descriptor for the initial K-NN search (\autoref{ssec:recognition}) as well as a set of local features to be deployed in the refinement step (\autoref{ssec:refinement}). For our test we choose kernel size equal $16$ and stride equals $2$ as to have 64 features per \textit{reference} image.

\subsection{Customer Use Case}
\label{ssec:customer_fine}

\begin{table}
	\centering
	\scalebox{0.9}{
		\begin{tabular}{|l|ccc|}
			\hline
			\textbf{Method}&\textbf{mAP(\%)}&\textbf{PR(\%)}&\textbf{mAMCA(\%)}\\
			\hline
			FV+RANSAC\cite{george2014recognizing}&11.26&23.14&6.41\\
			RF+PM+GA\cite{george2014recognizing}&23.49&43.13&21.19\\
			FM+HO\cite{yoruk2016efficient}&23.71&41.60&\textbf{32.50}\\
			\hline
			\textit{yolo\_gd} &21.49&47.03&13.34\\
			\textit{yolo\_ld }&27.84&53.17&16.32\\
			\textit{yolo\_ld+th}&30.46&37.88&28.74\\
			\textit{yolo\_ld+lf}&32.34&\textbf{58.41}&17.72\\
			\textit{yolo\_ld+mc}&30.15&55.25&21.09\\
			\hline
			\textit{yolo\_ld+lf-mc-th} \textbf{(full)}&\textbf{36.02}&57.07&31.57\\
			
			\hline
		\end{tabular}
	}
	\caption{Product recognition on the \emph{Grocery Products} dataset in the \textit{Customer} scenario. Best result highlighted in bold. Our proposals (in italic) yields large improvements in terms of both mAP and PR with respect to previously published results.}
	\label{tab:fine_customer}
\end{table}

In this sub-section we evaluate the effectiveness of our system in the \textit{Customer} scenario. To measure performance we rely on the annotations displayed in \autoref{fig:bboxes}-(a) and score a correct recognition when the product has been correctly identified and its bounding box has a non-empty intersection with that provided as ground-truth. We compare our method with already published work tested on the same dataset: FV+RANSAC (Fisher Vector classification re-ranked with RANSAC) \cite{george2014recognizing}, RF+PM+GA (Category prediction with Random Forests, dense Pixel Matching and Genetic Algorithm) \cite{george2014recognizing} and FM+HO (local feature matching optimized with Hough) \cite{yoruk2016efficient}. We report the results obtained by the different methods according to the tree metrics in \autoref{tab:fine_customer}. As \cite{yoruk2016efficient} does not provide the mAP figure but only the values of precision and recall, for FM+HO we report an approximate mAP computed by multiplying precision and recall. %For the other methods, instead, the mAP is computed as the approximate area under the Precision-Recall curve, where Precision and Recall are obtained by considering different confidence threshold on predictions.

%In the following, we will denote general purpose descriptors obtained from the publicly available VGG network trained on Imagenet with the suffix $\_gd$ (general descriptor), while  descriptors obtained from the network by fine-tuning according to the triplet loss based on \textit{Reference} images of  grocery products will be denoted by the suffix $\_ld$ (learned descriptor).

Using our trained \textit{yolo} network for product detection, in \autoref{tab:fine_customer} we report the results obtained by either deploying a general purpose VGG-based descriptor ($yolo_{gd}$) or learning an ad-hoc embedding for grocery products $yolo_{ld}$. Moreover, we report the results achieved with the different refinement strategies presented in \autoref{ssec:refinement}. 
\autoref{tab:fine_customer} shows that our pipeline can provide a higher recall than previous methods even with a general purpose image descriptor (\textit{yolo\_gd}), although with a somehow lower precision, as demonstrated by the slightly inferior mAP score. However, our complete proposal relies on learning an ad-hoc descriptor for grocery products (\textit{yolo\_ld}), which yields a significant performance improvement, as vouched by an average gain of about 6\% in terms of both Recall and mAP. Wrongly classified proposals can be discarded to further improve accuracy by the threshold refinement strategy (\textit{yolo\_ld + th} - with $\tau_d=0.9$), thereby increasing  the mAMCA from 16.32\% to 28.74\%. Re-ranking based on the proposed local features (\textit{yolo\_ld+lf}) turns out an effective approach to ameliorate both precision and recall, as demonstrated by a gain of about  5\%  in mAP and PR with respect to the pipeline without final refinement  (\textit{yolo\_ld}). The category-based re-ranking strategy (\textit{yolo\_ld + mc}) seems to fix some of the recognition mistakes and improve the recognition rate with respect to (\textit{yolo\_ld}) , providing gains in all metrics. Finally, by mixing all the refinement strategies to obtain our overall pipeline (\textit{yolo\_ld+lf-mc-th}), we neatly get the best trade-off between precision and recall,  as vouched by the 57.07\% PR and 36.02\% mAP, \ie about  14\% and 12.5\% better than previously published results, respectively, with a mAMCA turning out nearly on par with the best previous result.

%\subsection{Category recognition for customers}
%\label{ssec:customer_class}

\begin{figure}
	\centering
	\includegraphics[width=0.4\textwidth]{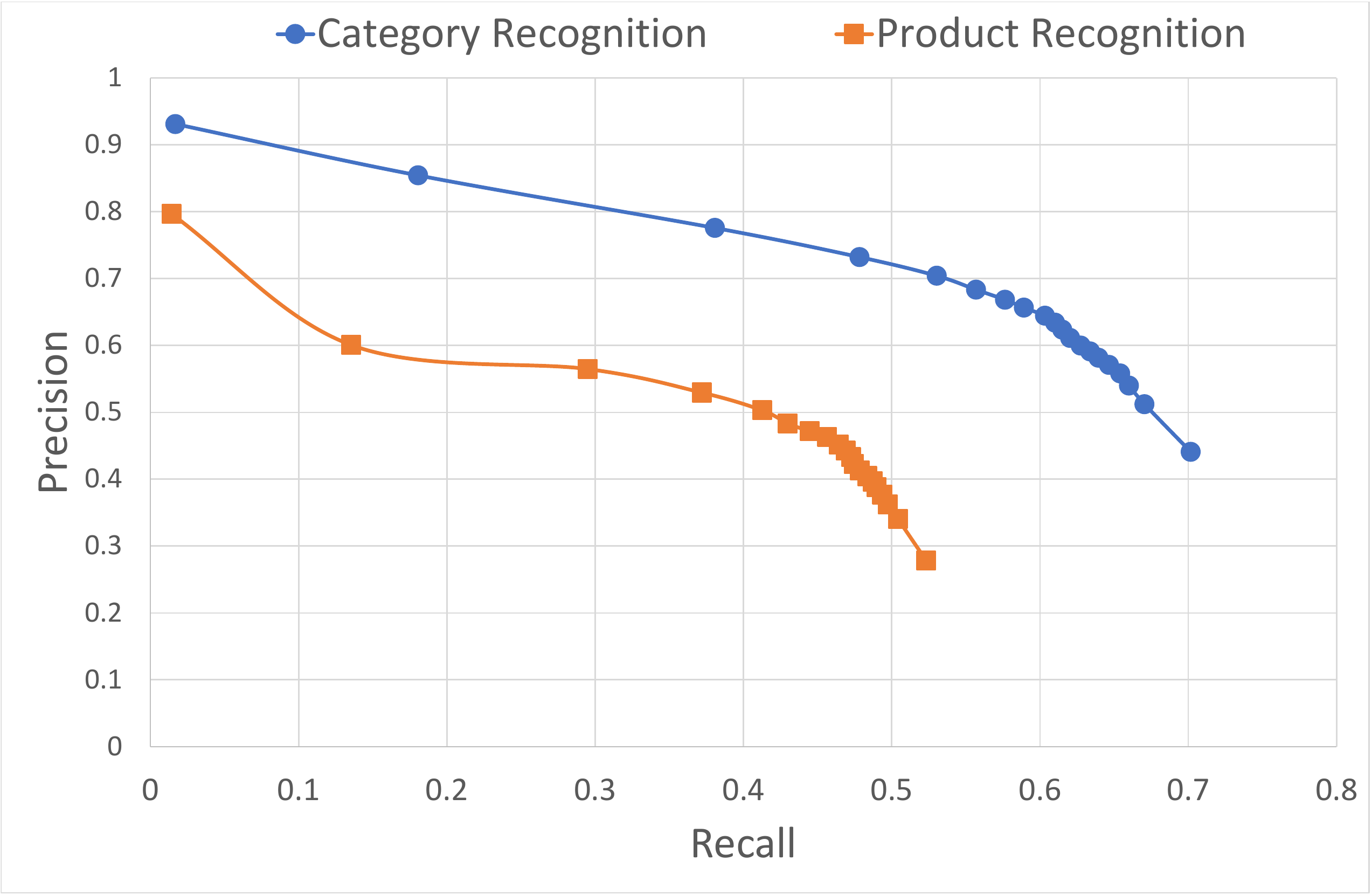}
	\caption{Precision-recall curves obtained in the \textit{customer} use case by the \textit{yolo\_ld} system when trying to recognize either the individual products or just their category.}
	\label{fig:catInstance}
\end{figure}

We found that casting recognition as a similarity search through learned global descriptors has the nice property that even when the 1-NN does not correspond to the right product, it usually corresponds to items belonging the correct category (\ie cereals, coffee,\dots). We believe this behavior being due to items belonging to the same category sharing similar peculiar visual patterns that are effectively captured by the descriptor and help to cluster nearby in descriptor space items belonging to the same categories (\eg, coffee cups often displayed on coffee packages or flowers on infusions). To highlight this generalization property, we perform here an additional test in the Customer scenario by considering a recognition as correct if the category of the 1-NN match is the same as those of the annotated bounding box. 
Accordingly, we compare the performance of \textit{yolo\_ld} when trying to recognize either the individual products or their category. The results of this experiment are reported as Precision-Recall curves in \autoref{fig:catInstance}. The large difference between the two curves proves that very often the system mistakes items at product level though correctly recognizing their category. %This property is obtained at zero cost given our problem formulation, however to enforce this behavior we could modify the training loss of the \textit{embedder} to force proximity for items belonging to the same category with a formulation similar to \cite{zhang2016embedding}. 
Eventually, it is worth pointing out that our method not only provides a significant performance improvement with respect to previously published results but turns out remarkably fast. Indeed, our whole pipeline can be run on a GPU in less than one second per image.

% For the sake of comparison,   \cite{yoruk2016efficient} reports a processing time of about 10 seconds per image.

\begin{table}
	\centering
	\scalebox{0.9}{
		\begin{tabular}{|c|cc|}
			\hline
			\textbf{Method}&\textbf{mAP(\%)}&\textbf{PR(\%)}\\
			\hline
			FS\cite{tonioni2017product}&66.37&75.0\\
			\hline
			\textit{yolo\_gd}&66.95&78.89\\
			\textit{yolo\_ld}&74.32&84.75\\
			\textit{yolo\_ld+th}&75.62&81.55\\
			\textit{yolo\_ld+lf}&76.37&\textbf{86.56}\\
			\textit{yolo\_ld+mc}&74.80&85.28\\
			\hline
			\textit{yolo\_ld+lf-mc-th} \textbf{(full)}&\textbf{76.93}&85.71\\
			\hline
		\end{tabular}
	}
	\caption{Product recognition for \textit{Management} use case. Our proposal highlighted (in italic), best results in bold.}
	\label{tab:fine180}
\end{table}

\begin{table}
	\centering
	\scalebox{0.9}{
		\begin{tabular}{|c|cc|}
			\hline
			\textbf{Method}&\textbf{mAP(\%)}&\textbf{PR(\%)}\\
			\hline
			FS\cite{tonioni2017product}&47.32&57.0\\
			\hline
			\textit{yolo\_gd}&60.17&73.66\\
			\textit{yolo\_ld}&67.88&80.27\\
			\textit{yolo\_ld+th}&69.70&76.01\\
			\textit{yolo\_ld+lf}&70.69&\textbf{82.83}\\
			\textit{yolo\_ld+mc}&69.01&81.55\\
			\hline
			\textit{yolo\_ld+lf-mc-th} \textbf{(full)}&\textbf{73.50}&82.66\\
			\hline
		\end{tabular}
	}
	\caption{Results in the \textit{Management} use case performing recognition against all the items belonging to the Food subclass of the \textit{Grocery Products} dataset ($\sim3200$). Our proposals highlighted (in italic), best results in bold.}
	\label{tab:miefull}
\end{table}

\begin{figure}
	\centering
	\includegraphics[width=0.4\textwidth]{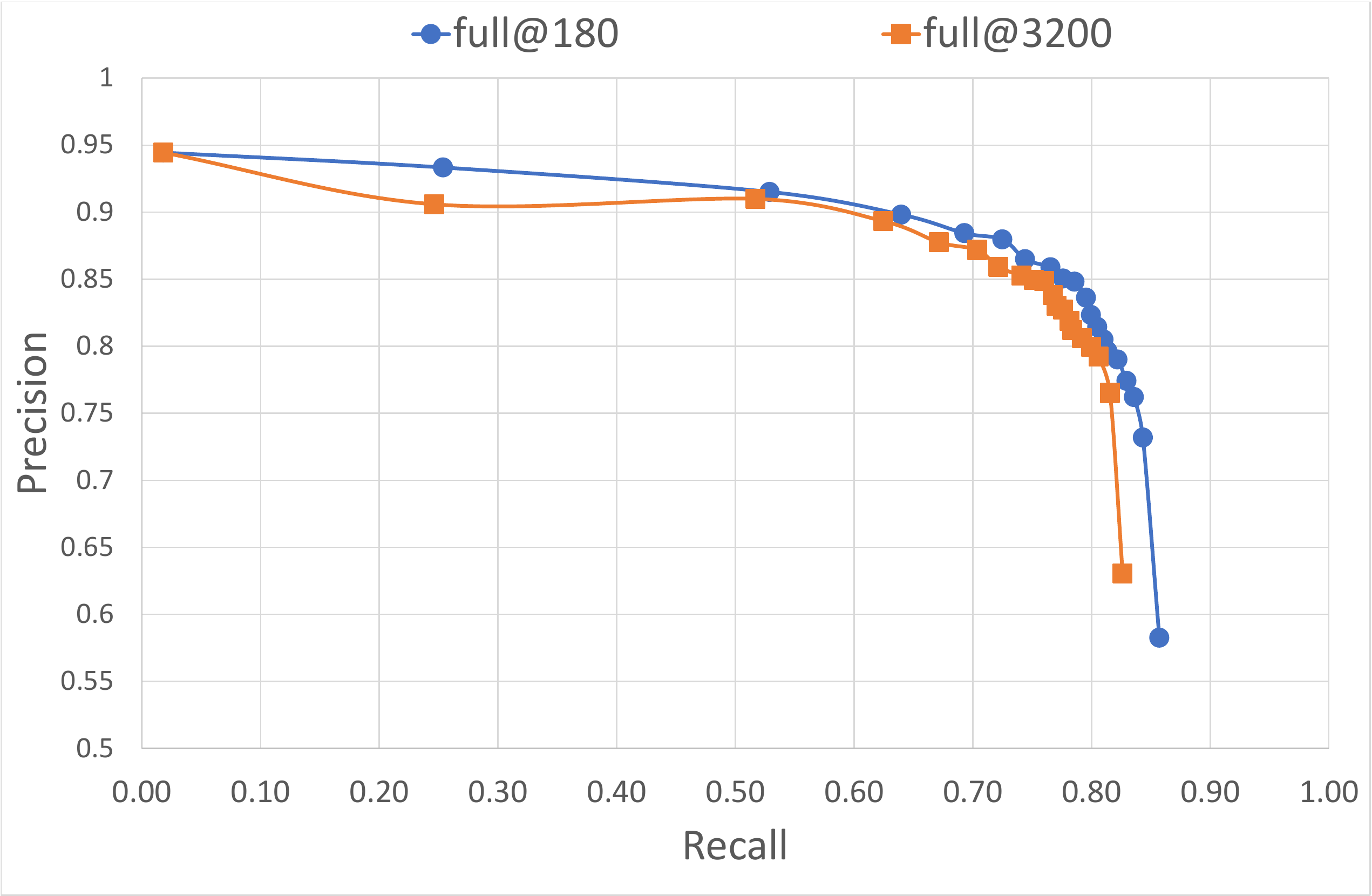}
	\caption{Precision-Recall curves for our full pipeline in the \textit{Management} use case. full@180 denotes performing recognition on the small \textit{reference} database of \cite{tonioni2017product} ($\sim 180$ entries), full@3200 against all the products in the \textit{Food} category of \emph{Grocery Products}  ($\sim 3200$).}
	\label{fig:zmiePrecRec}
\end{figure}

\subsection{Management Use Case}
\label{ssec:management_fine}
\begin{figure*}[t]
	\centering
	\begin{tabular}{ccc}
		\includegraphics[width=0.32\textwidth]{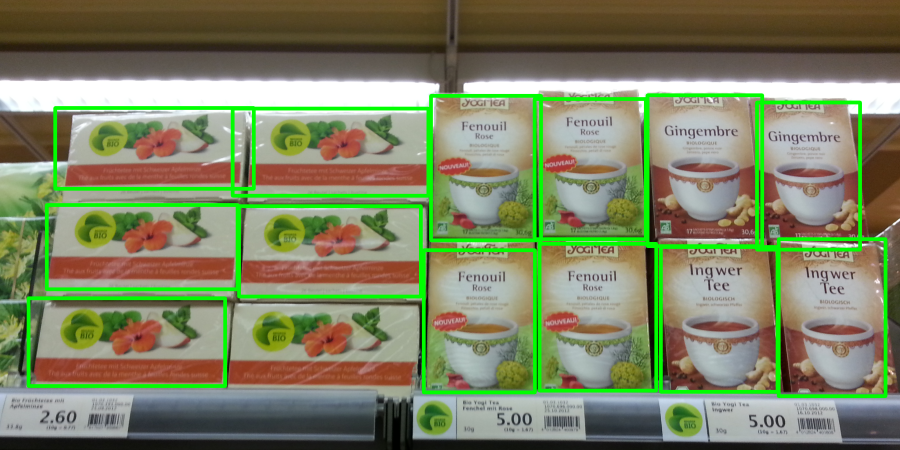}&\includegraphics[width=0.32\textwidth]{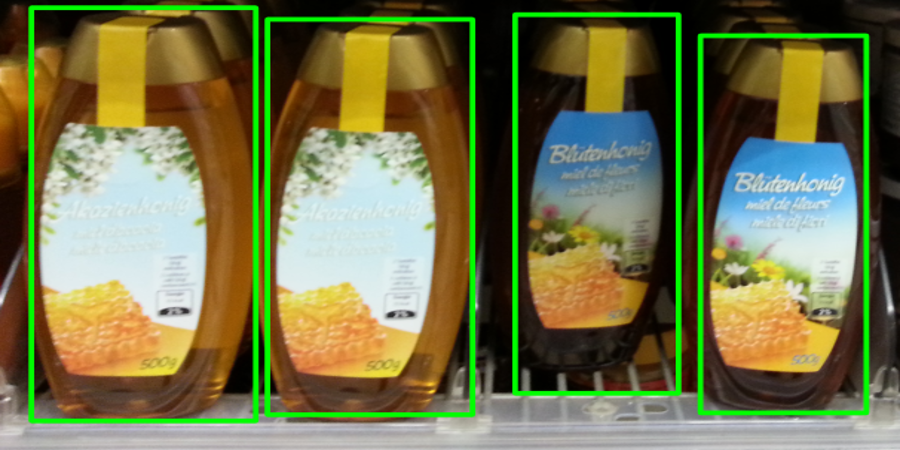}&\includegraphics[width=0.32\textwidth]{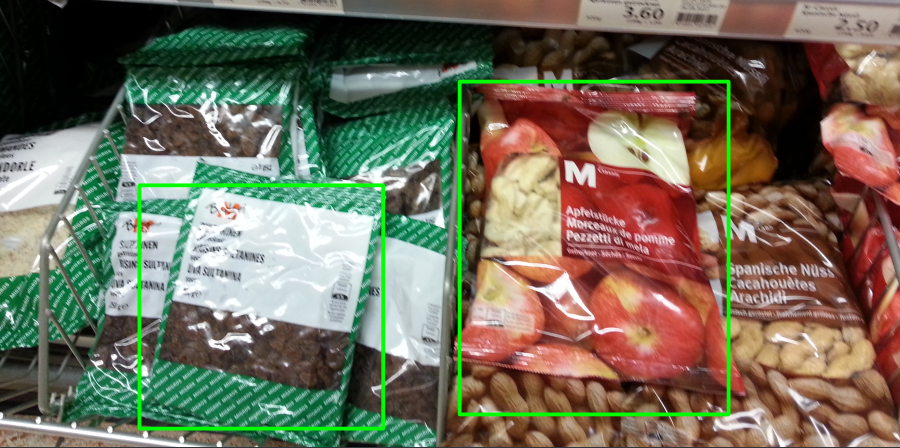}\\
		(a)&(b)&(c)\\
		%\includegraphics[width=0.3\textwidth]{qualita_4.png}\\
		%(c)&(d)\\
	\end{tabular}
	\caption{Examples of correct product recognitions in \emph{query} images from \emph{Grocery Products}.}
	\label{fig:qualitativi}
\end{figure*}

The experiments presented in this Section concern the \textit{Management} use case, a task  requiring correct recognition of all the individual products displayed on shelves. Thus, we rely on the annotations shown in \autoref{fig:bboxes}-(b) and consider a recognition as correct when the item has been correctly recognized and the intersection over union (IoU) between the predicted and ground truth bounding boxes is higher than 0.5. 
To the best of our knowledge, the only published results on the \emph{Grocery Products} dataset that deals with recognition of all the individual products are reported in \cite{tonioni2017product}. However, in  \cite{tonioni2017product} the authors address a specific task referred to as \textit{Planogram Compliance}, which consists in checking the compliance between the actual product disposition and the planned one.  Accordingly, the pipeline proposed in \cite{tonioni2017product} includes an initial unconstrained product recognition stage, which addresses the same settings as our \textit{Management} use case, followed by a second stage that deploys the exact knowledge on the planned product disposition in order to improve recognition accuracy and detect compliance issues (\eg, missing or misplaced products).
Therefore,  we compare our proposal to the most effective configuration of the first stage of the pipeline presented in \cite{tonioni2017product}, referred to hereinafter as \textit{FS}, which is based on matching BRISK local features\cite{leutenegger2011brisk} followed by Hough Voting and pose estimation through RANSAC.

%Given that for this specific task we are more concerned about the accurate localization o few items rather than having highly recall, we picked \textit{SSD} as \textit{detector} rather than \textit{yolo} as it produces less proposal, but more tightly placed around the products. 
To compare our pipeline with respect to \textit{FS}, we use the annotations provided by the authors and perform recognition against the smaller \textit{reference} database of 182 products used in \cite{tonioni2017product}. The results are reported in  \autoref{tab:fine180}. Firstly, it is worth pointing out how, despite the task being inherently more difficult than in the \textit{Customer} use case, we record higher recognition performance. We ascribe this mainly to the smaller subset of in-store images used for testing, (\ie, 70 vs. 680) as well as to these images featuring mainly rigid packaged products, which are easier to recognize than deformable packages. 
Once again, in our pipeline, the use of a learned descriptor (\textit{yolo\_ld}) provides a substantial performance gain with respect to a general purpose descriptor (\textit{yolo\_gd}), as the mAP improves from 66.95\% to 74.32\% and the PR from 78.89\% to 84.75\%.  The different refinement strategies provide advantages similar to those discussed in \autoref{ssec:customer_fine}, the best improvement yielded  by re-ranking recognitions based on the local features extracted from the \textit{Embedder} network (\textit{yolo\_ld+lf}). The optimal trade-off between precision and recall is achieved again by deploying together all the refinement strategies (\textit{yolo\_ld+lf-mc-th}), which provided a mAP and PR as high as 76.93\% and 84.75\%, respectively (\ie, both about 10\% better than the previously published results on this dataset).

\autoref{tab:miefull} reports results aimed at assessing the scalability of the methods with respect to the number of products in the \textit{reference} database. We carried out an additional experiment by performing the recognition of each item detected within the 70 \emph{query} images against all the 3200 products of the "Food" category in \emph{Grocery Products} rather than the smaller subset of 182 products proposed in  \cite{tonioni2017product}. 
By comparing the values in  \autoref{tab:fine180} and \autoref{tab:miefull}, we can observe how, unlike \textit{FS}, our method can scale nicely from few to thousands of different products: our full method \textit{yolo\_ld+lf-mc-th} looses only  3.43\% mAP upon scaling-up the \textit{reference} database quite significantly, whilst the performance drop for \textit{FS} is as relevant as  19.05\%. In \autoref{fig:zmiePrecRec} we also plot the precision-recall curves obtained by our full pipeline (\textit{yolo\_ld+lf-mc-th}) using the smaller  (full@180) and larger (full@3200) sets of reference products.  The curves show clearly how our pipeline can deliver almost the same performance in the two setups, which vouches for the ability of our proposal to scale smoothly to the recognition of thousands of different products. As far as recognition time is concerned, our pipeline can scale fairly well regardless of the size of the \textit{reference} database, due to the NN search, even if extensive, amounting to a negligible fraction of the overall computation: the difference in inference time between recognizing 180 and 3200 product is less than a tenth of a second. 

%both our proposal and \textit{FS} exhibit a complexity which increases with the number of items to be recognized. Yet, with our system only a single global descriptor must be added for each additional item, while pipelines based on local feature like \textit{FS} require adding thousands of local descriptors for each new item. 

\section{Qualitative Results}
\label{sec:qualitative}
\autoref{fig:qualitativi} reports some qualitative results obtained by our pipeline. Picture (a) shows the recognition results on an image taken quite far from the shelf and featuring a lot of different items; (b) deal with some successful recognitions in a close-up \emph{query} image,  where only a few items are visible at once. Finally, (c) refers to recognition of products featuring deformable and highly reflective packages, which are quite challenging to recognize due to the appearance of the items within the \emph{query} images turning out significantly different than in the available \textit{reference} images. Yet, in (c) our system was able to find at least one item for each product type (\ie, as required in the \textit{Customer} use case). %Additional qualitative results are provided in the supplementary material, which includes also some exemplary failure cases.

\section{Conclusion}
\label{sec:conclusion}
In this paper we have proposed a fast and effective approach to the problem of recognizing grocery products on store shelves. Our proposal addresses the task by three main steps: class agnostic object detection to identify the individual items appearing on a shelf image,  recognition through K-NN similarity search based on a global image descriptor, final refinement to further boost performance. All the three steps deploy modern deep learning techniques, as we detect items by a state-of-the-art CNN (\textit{Detector}), learn the image descriptor by another CNN trained to disentangle the appearance of grocery products (\textit{Embedder}) and extract local cues key to refinement as MAC features computed alongside with the global embedding.  

The experiments prove that our pipeline compares favourably to the state-of-the-art on the public dataset available for performance assessment while being remarkably fast. Yet, we plan to investigate how to further improve the speed at test time (\eg, to enable execution on a low-cost and/or mobile device). Purposely, we envisage devising a unified CNN architecture acting as both \textit{Detector} and \textit{Embedder}. Furthermore we are currently investigating on the use of generative models (\eg, GANs) to augment the number of samples per product to train the \textit{Embedder}. A generative model could also be trained to render \textit{reference} more similar to proposals cropped from \textit{query} images in order to shrink the gap between the training and testing domains.

\pagebreak

{\small
\bibliographystyle{ieee}
\bibliography{biblio.bib}
}

\end{document}